\documentclass[pdflatex,sn-mathphys-num]{sn-jnl}

\usepackage{graphicx}%
\usepackage{multirow}%
\usepackage{amsmath,amssymb,amsfonts}%
\usepackage{amsthm}%
\usepackage{mathrsfs}%
\usepackage[title]{appendix}%
\usepackage{xcolor}%
\usepackage{textcomp}%
\usepackage{manyfoot}%
\usepackage{booktabs}%
\usepackage{algorithm}%
\usepackage{algorithmicx}%
\usepackage{algpseudocode}%
\usepackage{listings}%
\usepackage{float}

\begin{document}

\title[Biologically-Informed Excitatory and Inhibitory Balance for Robust Spiking Neural Network Training]{Biologically-Informed Excitatory and Inhibitory Balance for Robust Spiking Neural Network Training}


\author[1]{\fnm{Joseph A.} \sur{Kilgore}}\email{jkilgore@gwmail.gwu.edu}

\author[2]{\fnm{Jeffrey D.} \sur{Kopsick}}\email{jkopsick@gmu.edu}

\author[2]{\fnm{Giorgio A.} \sur{Ascoli}}\email{ascoli@gmu.edu}

\author*[1]{\fnm{Gina C.} \sur{Adam}}\email{ginaadam@email.gwu.edu}


\affil*[1]{\orgdiv{Department of Electrical and Computer Engineering}, \orgname{George Washington University}, \orgaddress{\city{Washington}, \postcode{20052}, \state{DC}, \country{United States}}}

\affil[2]{\orgdiv{Center for Neural Informatics, Structures, \& Plasticity, Krasnow Institute for Advanced Study}, \orgname{George Mason University}, \orgaddress{\city{Fairfax}, \postcode{22030}, \state{VA}, \country{United States}}}


\abstract{Spiking neural networks drawing inspiration from biological constraints of the brain promise an energy-efficient paradigm for artificial intelligence. However, challenges exist in identifying guiding principles to train these networks in a robust fashion. In addition, training becomes an even more difficult problem when incorporating biological constraints of excitatory and inhibitory connections. In this work, we identify several key factors, such as low initial firing rates and diverse inhibitory spiking patterns,  that determine the overall ability to train spiking networks with various ratios of excitatory to inhibitory neurons on AI-relevant datasets. The results indicate networks with the biologically realistic 80:20 excitatory:inhibitory balance can reliably train at low activity levels and in noisy environments. Additionally, the Van Rossum distance, a measure of spike train synchrony, provides insight into the importance of inhibitory neurons to increase network robustness to noise. This work supports further biologically-informed large-scale networks and energy efficient hardware implementations.}




\maketitle

\section{Introduction}\label{sec1}
Current advances in artificial intelligence (AI) have led to increasingly power intensive training of large network models that is unsustainable in the long-term. These models are computed on farms of processor units and graphics cards, limited by inefficient data exchange between processing and memory storage \cite{efnushevaSurveyDifferentApproaches2017}. Due to these constraints, new methodologies are needed to fundamentally change the paradigm and envision efficient computation. The brain provides significant inspiration, due to its ultra-efficient capabilities of performing a wide range of tasks in an adaptive and robust fashion. Spiking neural networks (SNNs) inspired by the spiking behavior of the brain could therefore provide an energy efficient paradigm in machine learning \cite{pfeifferDeepLearningSpiking2018, NewNeuroAI2024}. Moreover, neuromorphic implementations of these spiking neural networks could support novel hardware accelerators, although more research is needed to understand how to train and map such biologically-realistic networks effectively. 

To further the understanding of training networks with progressively more biological realism, this work looks directly into the ratios of excitatory and inhibitory neurons within a network being trained. In biological networks representing the various regions of the mammalian brain, the ratio between excitatory and inhibitory neurons varies between approximately 10:90 to 90:10  with 80:20 being the approximate  excitatory:inhibitory (E:I) ratio across the adult mouse brain as a whole \cite{rodarieMethodEstimateCellular2022b}. Our work investigates networks with the ratios in these ranges relevant to rodent brain regions \cite{alrejaConstrainedBrainVolume2022} and analyzes their spiking activity. A range of network initial conditions were trained on two AI-relevant datasets, namely the Fashion-MNIST (Modified National Institute of Standards and Technology) image dataset and Spiking Heidelberg Digits (SHD) audio dataset. This testing gives a deeper look at the relationship between spiking behavior and network training performance. While previous work has relied on a mathematical derivation for optimal initialization, in this work we look for a more general activity-based solution that is potentially applicable beyond leaky-integrate-and-fire (LIF) models \cite{rossbroichFluctuationdrivenInitializationSpiking2022}. This solution provides some guidelines for increased training robustness and efficiency of relevance to spiking neural networks towards biologically realistic implementations.

This work aims to understand the importance and benefits of biologically-informed SNN design and analysis. To this end, the following key points summarize the contributions of this work:

\begin{itemize}
    \item Networks with sparser initial spiking activity, correlating to increased energy efficiency, successfully train more often than inefficient networks with higher initial firing rate.
    \item Networks with biologically realistic E:I ratios, e.g. 80:20 , train to higher accuracy in noisy training environments across different datasets.
    \item Analyzing SNNs with neuroscience-based metrics, e.g. the Van Rossum distance \cite{vanrossumNovelSpikeDistance2001}, provides insight into how different populations of neurons are statistically different from one another across training.
    \item In successfully trained, higher performing networks, and consistent with neurobiological evidence, inhibitory neurons display a greater diversity of spiking patterns than excitatory neurons, offering insight into how biologically-informed E:I ratios perform well in noisy environments.
    
\end{itemize}

\section{Results}\label{sec2}

The goal of this work is to understand the interplay of excitatory and inhibitory activity in networks trained on artificial intelligence tasks. To investigate this, we trained network architectures consisting of a spike generator input layer, a LIF neuron hidden layer with various E:I ratios, and a leaky-integrator output layer (Fig. 1a-b) . To expand on the understanding of excitatory and inhibitory neurons in a training environment, we divided the hidden layer into excitatory and inhibitory LIF neurons and tested a range of E:I ratios. The excitatory and inhibitory neurons are defined by their excitatory or inhibitory connections, with their weights to the following layer bounded either positive or negative, respectively. A LIF neuron is excitatory or inhibitory because a neuron itself has all excitatory or all inhibitory connections. The hidden layer consisted of 50:50, 80:20, 95:5, and 100:0 E:I neuron ratios. These ratios were chosen because they reflected the biological range with 80:20 corresponding to the whole brain, with 95:5 and 50:50 representing an extended range of E:I ratios observed in different brain regions \cite{rodarieMethodEstimateCellular2022b}. Additionally, 100:0 was tested for potential simplification of only excitatory neurons, simplification typically used in artificial neural networks typically used in the computer science literature. 

The resulting networks were tested on two representative datasets, the Fashion-MNIST dataset with latency encoded spikes and the SHD dataset \cite{xiaoFashionMNISTNovelImage2017, cramerHeidelbergSpikingData2022}. Each pixel of the Fashion-MNIST dataset was converted to a spike time, and 784 input spike generators corresponded to each of the 28x28 pixels. Training on Fashion-MNIST was conducted with a 100-neuron hidden layer. The SHD data has 700 neuron spike trains, which corresponded to 700 input spike generators, and utilized a 200-neuron hidden layer. The output leaky integrators allowed for network activity to be aggregated into 10 or 20 neurons corresponding to the dataset classes of Fashion-MNIST and SHD, respectively. The network initialization involved generating connections using normal distributions of varying standard deviations and bounding the weights to either excitatory (positive) or inhibitory (negative).  Networks were trained for 30 epochs on the Fashion-MNIST dataset, and 200 epochs on the SHD dataset. After training, multiple metrics were employed including classification accuracy, weight distributions, as well as spike frequency, proportional E:I activity, and Van Rossum distances. All networks were trained using the surrogate gradient method \cite{neftciSurrogateGradientLearning2019}. Additional details regarding network architecture, training and testing protocols, and network analysis can be found in the methodology section.

\subsection{Network Accuracy vs Initial Firing Rate}
The initial networks trained on Fashion-MNIST were able to train to over 80\% accuracy, which is similar to networks of the same size with unbounded weights \cite{neftciSurrogateGradientLearning2019} (Fig. 1c).  For networks with E:I ratios 80:20, 95:5, and 100:0, the best performing networks were initialized within the biologically-realistic bounds between 0.01 Hz and 25.6 Hz \cite{sanchez-aguileraUpdateHippocampomeOrg2021}, leading to an energy-efficient implementation \cite{harrisSynapticEnergyUse2012} while still retaining a high accuracy performance. Networks with a high percentage of inhibitory neurons, e.g. 50:50, were able to train at the lowest end of the tested activity range, but the accuracy performance was not robust across repeat trials. In contrast, networks without inhibitory neurons exhibited high accuracy over a larger range of initial activity levels. However, these upper activity levels would not be energy efficient due to unnecessarily high spiking activity levels, nor do they confer additional accuracy benefits. Therefore, we focused on the lower activity spectrum for further analysis. 

The SHD accuracy results mirror a similar trend with respect to activity levels and accuracy (Fig. 1d). Lower activity levels correspond to higher accuracy across all E:I ratios. Accuracies reached ~45\%, which was comparable to the accuracy of the unbounded hidden layer for this network architecture \cite{cramerHeidelbergSpikingData2022}. An interesting result is that the purely excitatory network trained on SHD show poor accuracy, probably because of the intrinsic high level of input noise in this dataset. The networks with inhibitory neurons train to higher accuracy, which is an indication of the importance of excitatory:inhibitory balance in noisy environment. Taken together, networks with biologically realistic E:I of 80:20 maximized accuracy on both datasets when initialized with low levels of activity.

\begin{figure}[t]
\centering
\includegraphics[width=0.9\textwidth]{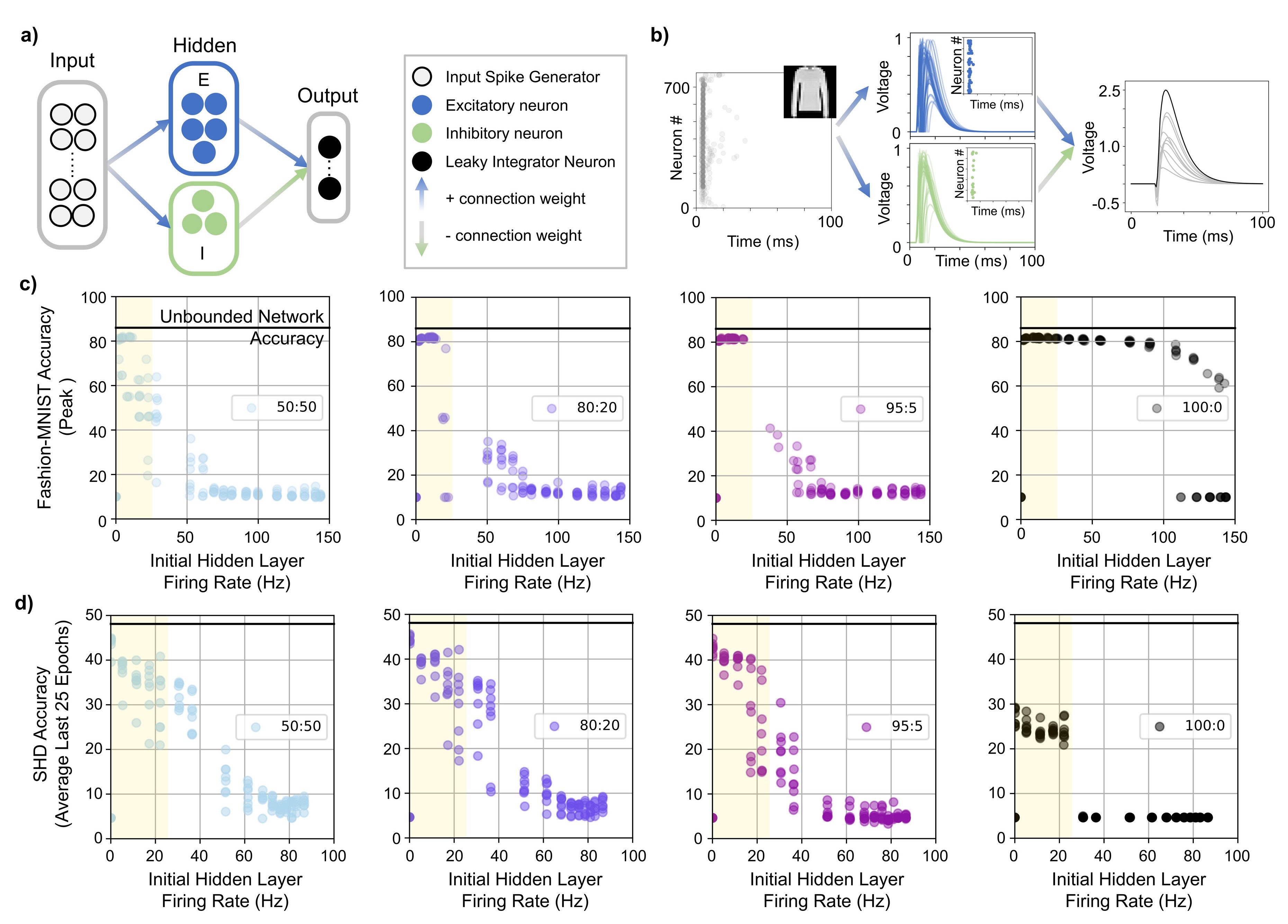}
\caption{Network protocol overview and training results. a) The spiking networks consist of three unique fully connected layers. The input layer consists of excitatory spike generators. The hidden layer consists of LIF neurons that have either excitatory or inhibitory weight connections to the output layer of leaky-integrators. b) Sample Fashion-MNIST activity of the network starting with the input layer. Spikes of the hidden layer are indicated with a raster plot inset. Hidden layer activity is then propagated to the output where the highest voltage trace corresponds to the networks chosen class (drawn in black). c) Network ability to train on the Fashion-MNIST dataset relative to their initial hidden layer firing rate. The different activity levels correspond to the different initial weight distributions. Every trial corresponds to an individual point on its respective graph. Training consisted of 4 different E:I ratios (left to right 50:50, 80:20, 95:5, 100:0). The black line at 86\% accuracy shows the accuracy of previous training of a network of the same size but with unbounded weights. d) Network ability to train on the SHD dataset across a range of initial hidden layer firing rates. From left to right 4 E:I ratios: 50:50, 80:20, 95:5, 100:0. The black line at 48\% accuracy represents a network of the same size but with unbounded weights.}\label{fig1}
\end{figure}

\begin{figure}[t]
    \centering
    \includegraphics[width=0.9\linewidth]{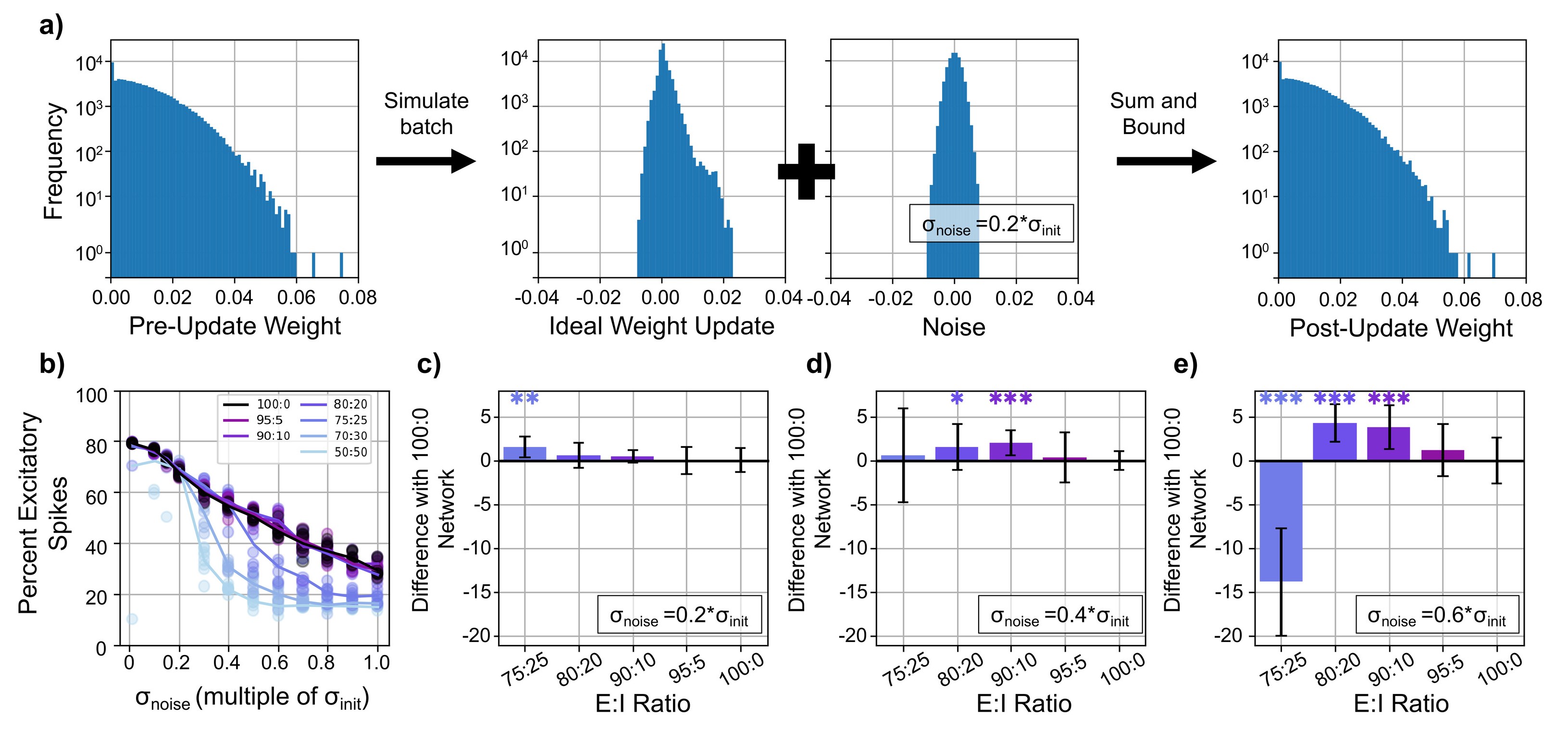}
    \caption{Accuracy of networks with noisy weight updates. a) Weight updates where the pre-update distribution determined the gradient for the update. The gradient and noise distributions were then combined with the weight distribution (this example was an $\sigma_{noise}=0.2\sigma_{init}$) and bounded to get the post-update weight distribution. b) Average accuracy of the final 10 of 30 epochs of training with 16 repeat trials for each E:I ratio and noise level combinations. The average across repeat trials is represented with corresponding lines. Each trial is represented with individual points. Average accuracy across trials compared to the 100:0 E:I with positive accuracy differences indicating an improvement over the 100:0 networks at three different $\sigma_{noise}$ levels: (c) $0.2\sigma_{init}$, (d) $0.4\sigma_{init}$, and (e) $0.6\sigma_{init}$. T-tests were performed to calculate the difference between each E:I ratio and the baseline 100:0 network with * indicating p $<$ 0.05, ** for p $<$ 0.01, and *** for p $<$ 0.001.}
    \label{fig:noisy-accuracy}
\end{figure}

\subsection{E:I Ratio vs. Robustness to Noisy Weight Updates}
To test the potential for mapping such excitatory:inhibitory balanced networks to noisy hardware, noisy weight updates were implemented by adding a normal distribution of varying standard deviations at every weight update. Each update included a calculated gradient and random noise (defined by $\sigma_{noise}$) that were added as a percentage of the initial weight distribution (defined by $\sigma_{init}$), which was then bounded to the excitatory or inhibitory values (Fig. 2a). Emerging device technologies, such as oxide-based resistive switches, experience high levels of stochasticity in the weight updates particularly in the high resistance state, therefore we have modeled a broad range $\sigma_{noise}$ from 0\% to 100\% of $\sigma_{init}$ \cite{grossiFundamentalVariabilityLimits2016}. Experimental results\cite{garbinVariabilityStudyPCM2015} support this range which corresponds to the experimental stochasticity of oxide-based resistive switches in the range of 1\% to 100\% of the initial resistive states $10^3 \Omega$ to  $10^6 \Omega$. Training was conducted over a range of E:I ratios initialized with low activity and trained on the Fashion-MNIST dataset. As the noise level increased, the accuracy decreased linearly for higher E:I ratios. Comparing the accuracy across ratios, the lower ratios (e.g. 70:30) performed at higher than 100:0 accuracy until higher noise levels where accuracy dropped off to near random chance (10\%) (Fig. 2b). For example, the 50:50 network performance dropped at $\sigma_{noise}$ equal to 30\% of $\sigma_{init}$, ($\sigma_{noise} =0.3 \sigma_{init}$) while other networks performed at 60\% accuracy.

To foster a comparison with 100:0 networks typically used in traditional neural networks, we assessed average accuracies across E:I ratios at three different noise levels (Fig. 2c-e). At the lower noise range, $\sigma_{noise} =0.2 \sigma_{init}$, networks performed near 70\% accuracy (Fig. 2c). Comparing the 16 repeat trials with a t-test at each E:I ratio, the 75:25 network performed at a higher accuracy than the 100:0 network at $\sigma_{noise} =0.2 \sigma_{init}$ ($p < 0.01$). As the noise level increased, the 80:20 and 90:10 networks were the highest performing networks (Fig. 2d-e). While previous noiseless Fashion-MNIST results observed the 100:0 network was robust over a larger range of initial activity, these results indicate a balance of excitatory and inhibitory activity provided optimal performance in the presence of noise.

\subsection{Successful vs. Unsuccessful Training}
To further understand the network training, the excitatory vs. inhibitory spiking activity can be observed by epoch for all classes to observe any differences between successful and unsuccessful networks. Activity levels before, during, and after training were visualized for the Fashion-MNIST dataset (Fig. 3). Initial activity spans from less than one hidden layer spike per image on average to over 1000 spikes per image. The initial activity falls along the ratio of the initial networks, 95\%, 80\%, and 50\%. After a single epoch of training, the separation between successful networks (overall accuracy over 50\%) was visible across the activity levels (Fig. 3a). Additionally, the networks that increased in accuracy have a wider range of percent excitatory activity.

By comparison, networks that failed to train remain close to their original E:I ratios (Fig. 3b). While some individual classes of a network increased in accuracy (with an increased percent of excitatory activity), other classes sustained low accuracy (with a decreased percent of excitatory activity). The networks that have high initial activity were unable to train regardless of the excitatory activity percentage. Overall, this provides further insight on why some networks successfully train on Fashion-MNIST while others do not. In addition, the best training occurs when the percent of excitatory activity increases from baseline, as seen in both the successful networks, and the classes that trained in the failed networks.

\begin{figure}[t]
    \centering
    \includegraphics[width=0.9\linewidth]{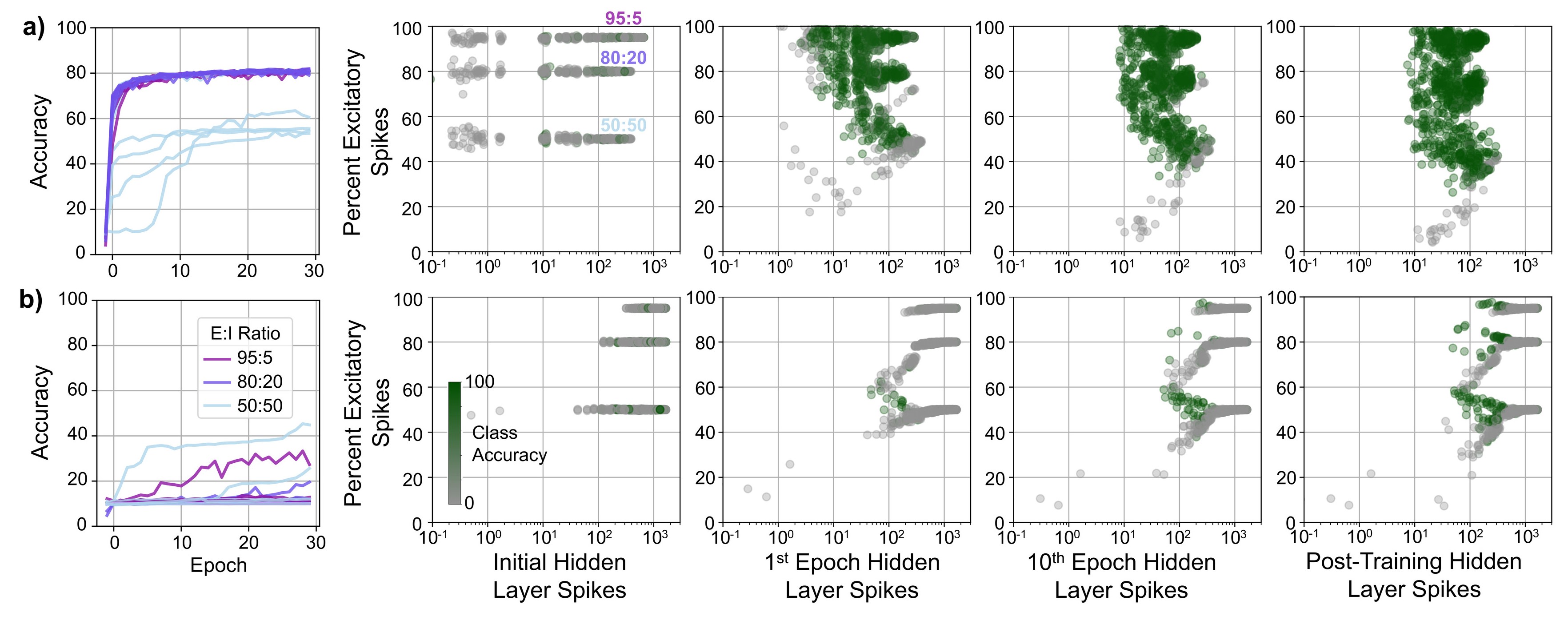}
    \caption{Fashion-MNIST class accuracy relative to the amount of activity and percent of activity that is excitatory across training. Separation is observed between (a) high accuracy networks ($>$50\% maximum overall accuracy) at lower activity levels, and (b) low accuracy networks at high activity levels (first column). (Second to fifth column) The initial network, after a single epoch of training, after ten epochs of training, and after training is complete (thirty epochs). Each network is represented by 10 points (for the 10 classes of the dataset) and all networks tested from Figure 1 with E:I ratios of 50:50, 80:20, and 95:5.
}
    \label{fig:class-accuracy-and-activity}
\end{figure}

\subsection{Evolution of E:I Activity During Training}
The interplay of excitatory and inhibitory  activity across Fashion-MNIST training classes was further analyzed for three representative trials of low, moderate, and high initial activity, all of which can train to over 50\% accuracy (Fig. 4). Although even higher initial activity trials exist, those do not successfully train (see Fig. 1 and 3 for the full activity range). In the low initial activity networks, the first epoch was categorized by a large increase in activity (Fig. 4a). Additionally, the increased activity was disproportionately increased toward excitation, shown in the increased percent of excitatory activity in the first epoch of training. The moderate initial activity trials also displayed initial increased excitation, even though the overall activity remained relatively constant (Fig. 4b). The remainder of training for the low and moderate activity had a continuous incremental increase in inhibitory activity while the excitatory activity remained constant or slowly declined. 

This trend was also consistent among classes within the dataset. Due to the conversion process from pixels to the spiking domain, objects that took up larger portions of the 28x28 pixel space generated more network activity. This was observed with the pullover class (labeled 2) having generated over 10 times the hidden layer activity at the end of training compared to the sandal class (labeled 5). The ratio of excitatory and inhibitory activity of lower activity classes tended to be higher than that of higher activity classes. 

The example high initial activity shows an initial reduction in the percent of excitatory activity (Fig. 4c). The accuracy also does not increase until near epoch 10, at which time the proportion of excitatory activity for some classes  increases. In this case, the classes that increased in proportional activity were the lower activity sandal and sneaker classes, labeled classes 5 and 7, respectively. Progressively as the training continued, similar to the other initial conditions, the excitatory activity decreased, but so did the inhibitory activity.

These general trends for low, moderate, and high initial activity are consistent throughout different initial E:I ratios. It should be noted that for higher initial ratios, the initial boost in percentage of excitatory activity was not visible since the activity was already at a percentage higher than that seen in the increases observed in the 50:50 trials. Therefore, the 50:50 trials were included here, as the other trends seen in the other E:I ratios are also seen in the 50:50 example. Additional examples for the 80:20 and 95:5 ratios are provided in Supplemental Fig. 2 and 3.

\begin{figure}[H]
    \centering
    \includegraphics[width=0.7\linewidth]{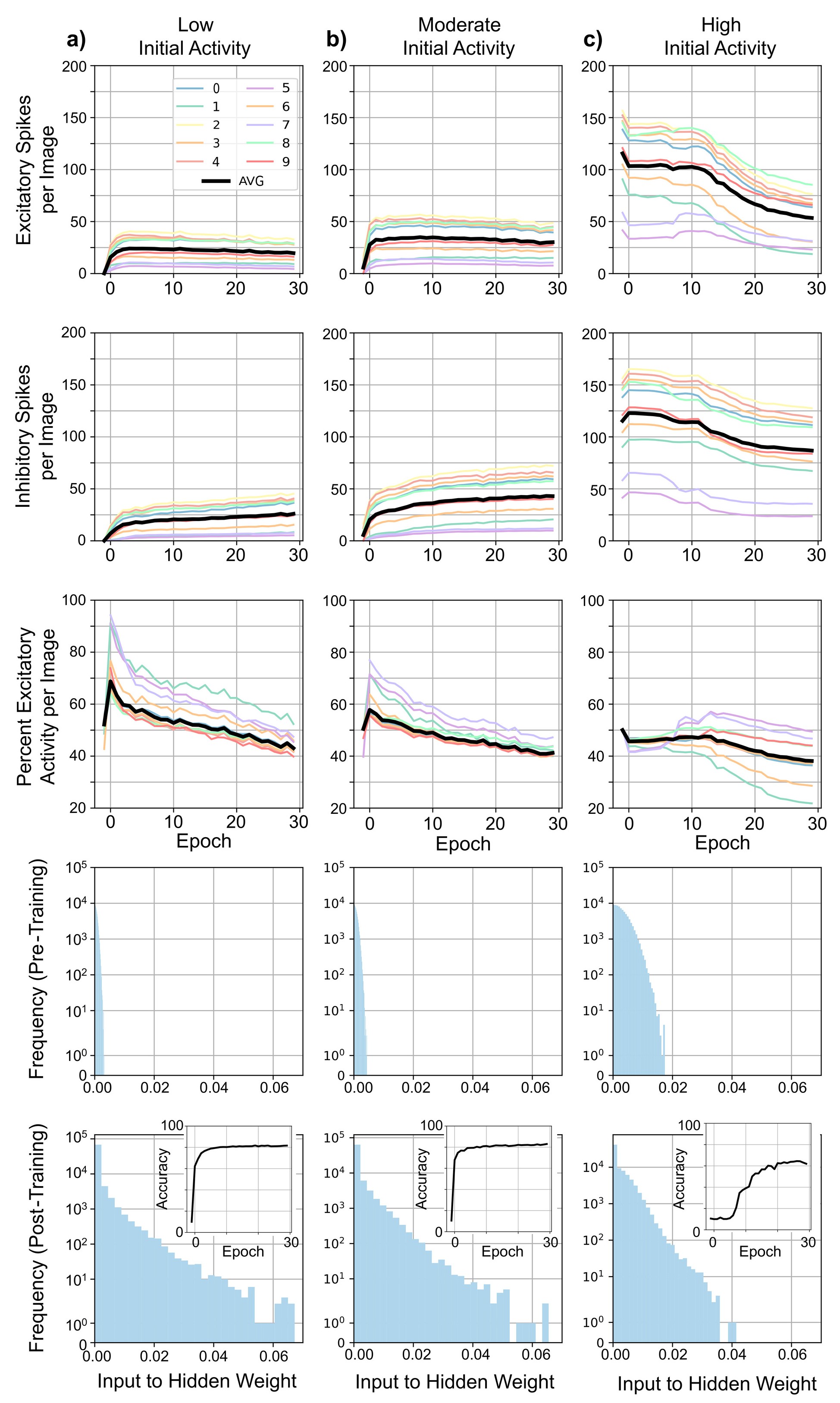}
    \caption{Representative activity of three 50:50 Fashion-MNIST trials. The examples cover (a) low initial activity, (b) moderate initial activity, and (c) higher initial activity. The average number of excitatory (first row) and inhibitory (second row) spikes per image denoting the number of spikes generated by the hidden layer on average given any input image in the Fashion-MNIST dataset. Each line represents a different class with the black line noting the average across all classes. The percentage of excitatory spikes across training (third row) indicates the initial increase in proportional excitatory activity, followed by a decrease for the remainder of training in the low and moderate initial activity conditions. The adjustment of activity from the network training perspective, derived from the weight distributions before (fourth row) and after (fifth row) training. The accuracy convergence curve of each individual trial is included as the inset of the post-training weights. 
}
    \label{fig:activity-training}
\end{figure}

\subsection{Distances Between Neuronal Pairs}
To further contrast successfully and unsuccessfully trained networks, the Van Rossum distance, a measure of dissimilarity between spike trains, was calculated between pairs of excitatory-excitatory (E-E), excitatory-inhibitory (E-I), and inhibitory-inhibitory (I-I) neurons for the Fashion-MNIST and SHD datasets (Fig. 5). By taking the average distance between all pairs across the entire dataset, all trials can be analyzed to compare between the three pair categories. Initially, the distribution of distances was the same regardless of initial excitatory and inhibitory ratio or pair category. The post-training trials are again divided between successful and failed networks. Fashion-MNIST success was defined as peak accuracy being greater than 50\%. SHD success was defined as the average accuracy of the last 25 epochs being greater than 30\%. 

For the Fashion-MNIST distance, successful networks have a lower range of distances compared to the failed networks, corresponding with the overall activity level of the networks since higher activity levels correspond to higher distances (Fig. 5a-c). Successful 80:20 networks had an E-E median distance of 0.890 while failed networks were significantly higher at 1.532 median E-E distance. Additionally, we observed that the categories of distances in successful 50:50 networks were significantly different (p = 0.0091 Kruskal-Wallis, Table 2). Interestingly, increased E:I ratios of 80:20 and 95:5 displayed further disparity in distances between the three different categories for successful networks (p $<$ 0.0001, Kruskal-Wallis, Table 2). The median Van Rossum distance between E-E neuron pairs was less than that of the I-I neuron pairs for all E:I ratios , with the 50:50 network showing the greatest increase of 39.6\% between E-E and I-I median distance. This indicates the activity of inhibitory neurons are further differentiated compared to the excitatory neurons, which is in line with biological observations.

The trend across pair categories within measurements from the SHD dataset remains the same as the Fashion-MNIST dataset with E-E distances being the lowest average distance (Fig. 5d-f). Additionally, the I-I distances have a higher median compared to the E-E pairs (see Table 3 for the Kruskal-Wallis statistical analysis). This further indicates the importance of differentiation of inhibitory neural activity in successful networks. In contrast, the range of distances of successful networks after being trained was higher than the initial range of distances, showing a greater increase in distances between initial and final networks on the SHD dataset. Furthermore, the range of distances from successful SHD dataset training were above 1.5, while Fashion-MNIST distances ranged below 1.5, indicating increased complexity of the SHD dataset and the increased need for differentiation of neurons to classify more complex data.

\begin{figure}[t]
    \centering
    \includegraphics[width=0.9\linewidth]{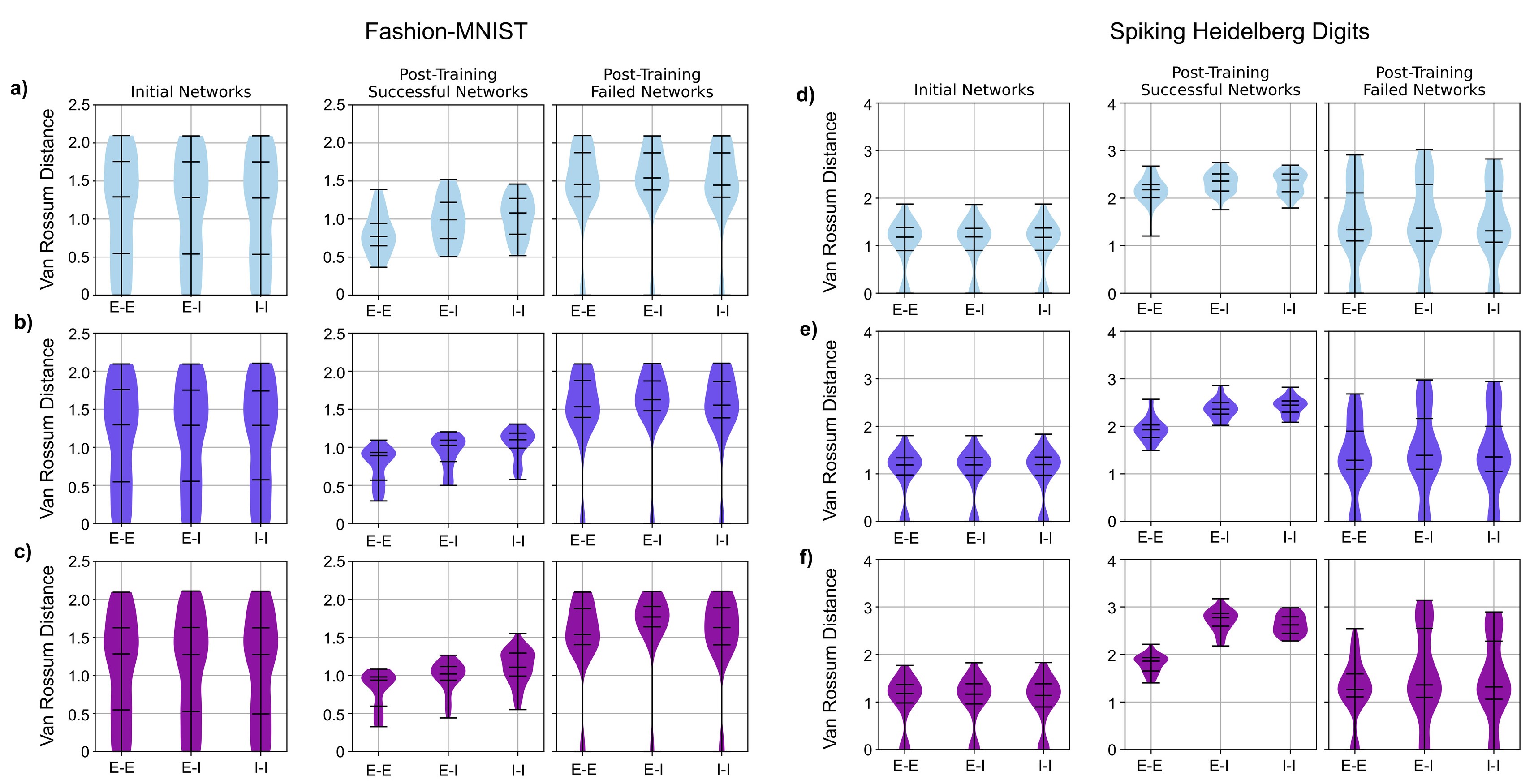}
    \caption{Average Van Rossum distances between hidden layer neuron pairs trained on the Fashion-MNIST (a-c) and SHD (d-f) datasets. Rows correspond to different E:I ratios 50:50 (a,d), 80:20 (b,e), 95:5 (c,f). The average distances between neuron pairs: excitatory-excitatory (E-E), excitatory-inhibitory (E-I), and the inhibitory-inhibitory (I-I) pairs. The distance is averaged across all cases. Each network tested in Figure 1 is represented by a single average value for each of the three categories and make the distributions in each subfigure. The initial networks spread a range of distances based on their varied weight initializations but show equal distributions across the three categories. The successfully trained networks from the Fashion-MNIST trials are defined as networks with peak overall accuracy $>$50\%, and have a lower distribution of distances, while the networks that fail to reach 50\% accuracy are typically in a higher range of activity, with the difference between failure and success being stronger in higher E:I ratios. Successfully trained networks from the SHD trials are defined as networks with average accuracy $>$30\% for the last 25 epochs. Using the Kruskal-Wallis test, the three categories are calculated to be statistically significantly different with p-values $<$0.0001 for the successful 80:20 and 95:5 networks, with I-I distances being the largest, and E-E distances being the lowest.
}
    \label{fig:van-rossum-distances}
\end{figure}

\section{Discussion}\label{sec12}
Overall, our results demonstrate that a biologically realistic range of E:I ratios between 80:20 and 90:10 in the hidden layer works best in the energy efficient activity range as well as being robust in noisy environments. In all four E:I ratios, we observed trials successfully training to $>$80\% accuracy on the Fashion-MNIST dataset, and $>$40\% accuracy on the SHD dataset. Both results are near the benchmark accuracy level of an equivalent network with unbounded weights. We observed that the accuracy of networks at low activity is similar across E:I ratios for the Fashion-MNIST image dataset, but the inclusion of inhibitory neurons in our networks provide significant accuracy benefits on the noisy SHD audio dataset. Additionally, the inclusion of noise within weights in the Fashion-MNIST training finds that the inclusion of inhibitory neurons is important for network robustness. Future work will investigate the potential of transmodal training for these networks using audio-visual datasets.

Our results also indicate that as network activity increases, the overall peak performance of the network decreases, eventually leading to 10\% accuracy (the baseline accuracy of the Fashion-MNIST dataset). Based on this data, we hypothesize that the increased initial activity was too high to provide any substantial differentiation between classes, and thus reduces the ability for the network to train. When the minimal spiking activity occurs, the activity is at an optimal point of differentiability and trains to a higher degree of accuracy. This trend remains across all E:I ratios, though the rate at which accuracy decreases as activity increases varies across E:I ratios.

To determine the E:I ratio optimal for potential translation to a neuromorphic hardware implementation, we utilized a noisy environment with comparable noise levels in the weight updates to compare the networks in the low activity range. From these trials, we see that networks within the biologically realistic E:I ratios (below 95:5) allow for the best performance in terms of accuracy within the noisy environment with $\sigma_{noise}$ between 20\% to 60\% of the $\sigma_{init}$. This indicates that the biologically realistic range of E:I ratios also provides the most benefits in combination with the biologically realistic firing range in hardware realizations.

On a macroscopic level, lower activity while being the most energy efficient is optimal for maximizing accuracy. The three representative activity levels see the trends toward a central activity level (approximately 50-100 hidden layer spikes). The other key feature of the successfully trained networks seems to be the immediate increase in excitatory activity in the first few epochs. This would correspond with the network needing excitatory activity to create a general picture of each class, and then the progressive increase in excitatory activity helps specify that image. This is reflected in the Van Rossum distances with successful networks having a low E-E distance, indicating the similarity between the spike trains of excitatory neurons.

Conversely, the inhibitory neurons had the greatest average distance between pairs, noting their differentiation of activity. This result is in line with experimental results from neuroscience, where biological networks are known to have greater physiological diversity across interneurons. Inhibitory interneurons exhibit differentiation in their connectivity \cite{fishellInterneuronTypesAttractors2020}, spiking behavior \cite{komendantovQuantitativeFiringPattern2019}, and down to the genetics of each interneuron type \cite{wamsleyGeneticActivitydependentMechanisms2017}. In our network after training, even though the LIF models themselves are identical, the weights of the network support an emergent diversity in inhibitory activity. Future work that incorporates biologically-realistic neuronal models would likely provide even more spiking diversity across training furthering the need for biologically-informed E:I ratios.

Previously, other work has expressed methods for network initialization using a mathematical approach \cite{rossbroichFluctuationdrivenInitializationSpiking2022}. Our work separates from the traditional mathematics of artificial SNNs in favor of an analysis utilizing metrics from neuroscience. This biologically-inspired approach  could also more readily provide insight into other networks beyond the LIF models, and potentially towards more complex models \cite{izhikevichDynamicalSystemsNeuroscience2007a, venkadeshSimpleModelsQuantitative2019, teeterGeneralizedLeakyIntegrateandfire2018}.   It should also be noted that this work separates from the traditional neuroscience, where prior works have focused the stability of  large biologically-realistic models in terms of spiking activity \cite{gastNeuralHeterogeneityControls2024, deneveEfficientCodesBalanced2016}, ours focuses on the actual use of network variants in AI-relevant classification tasks. From the AI perspective, previous work has sought to simplify the initialization of large-scale artificial neural networks \cite{dingParameterefficientFinetuningLargescale2023}. Based on our results we find key guiding principles, such as biologically-informed E:I ratios, that can be applicable to SNNs. In these comparisons, our work seeks to bridge the gap between artificial intelligence and neuroscience. 

In this work, we investigated the network conditions conducive to successful training via surrogate gradient backpropagation. Our results demonstrate that the minimal spiking activity produces networks that train successfully. We also observe that the optimal balance of stability in noisy datasets and noisy training environments are best met in a biologically realistic E:I ratio, thus indicating that balanced E:I activity would likely be required for successful neuromorphic hardware based on imperfect devices. Potentially, large biologically realistic networks can be initialized based on this minimal activity principle and biologically-informed parameters to avoid an exhaustive and computationally expensive parameter search. Furthermore, these results allows for future work to continue towards biological realism with modified detailed neuron models and data-driven structural diversity \cite{wheelerHippocampomeOrgKnowledge2024, kopsickRobustRestingStateDynamics2023}  as well as providing insight for hardware implementations.

\section{Methods}\label{sec11}
\label{sec:methods}
\subsection{Network Architecture and Training Protocol}
The network architectures were feedforward networks consisting of three layers (Fig. 1a). The size of the network was dependent on the dataset with the Fashion-MNIST dataset having sizes of 784-100-10 for the input-hidden-output layers. The SHD dataset training utilized networks of size 700-200-20. Each layer was fully connected to the following layer and initialized with normally distributed weights. Simulations lasted for 100 ms with 1 ms time steps for the Fashion-MNIST data and 200 ms for the SHD data  and calculated using the forward Euler method (see Fig. 1). The following equations describe the synaptic current and LIF model used: 
\begin{equation}
I_{syn}[t+1] = e^{-\frac{1}{\tau_{syn}}}I_{syn}[t] + \sum_i (W_i S_i[t])  
\end{equation}
\begin{equation}
V[t+1] = e^{-\frac{1}{\tau_{mem}}}V[t] + I_{syn}[t]   
\end{equation}

\begin{equation}
if V[t]>1 \longrightarrow V[t+1] = 0, S[t]=1
\end{equation}

Where $I_{syn}[t]$ is the synaptic current that decays with a $\tau_{syn}$ = 5 ms. Each time step the current spikes, denoted with $S_i[t]$ from each of the $i$ neurons in the previous layer were multiplied by their connecting weight,  $W_i$, which generated an increase in synaptic current. The voltage of each hidden layer neuron, $V[t]$, had an exponential decay with $\tau_{mem}$ = 10 ms. Finally, if the voltage was greater than 1 mV, the neuron would fire and reset to 0 mV. A negative log-likelihood loss function was calculated at each time step and used with an exponential surrogate gradient function to calculate weight updates for training \cite{neftciSurrogateGradientLearning2019}. Simulations were repeated with a batch size of 256 for 30 epochs of training on the Fashion-MNIST dataset and 200 epochs of training on the SHD dataset.

The hidden layer was constructed of excitatory and inhibitory neurons. Excitatory and inhibitory neurons were implemented with weights that were bounded after each update using a rectified linear unit (ReLU)  function  . This caused all weights that changed sign after an update to be set to 0, and all other weights left unaffected. Initialization of connections were randomized using the absolute value of a normal distribution for weights coming from an excitatory neuron, and the negative absolute value of a normal distribution for weights coming from an inhibitory neuron. The input layer consisted entirely of excitatory neurons to create activity in the subsequent layer of the network.

To analyze the network activity, a range of input to hidden network connections was tested. The input to hidden connections were generated using the absolute value of a normal distribution using a range of standard deviations (changing the initial hidden layer firing rate). Each standard deviation was applied on four E:I ratios : 50:50, 80:20, 95:5 and 100:0 with eight repeat trials and trained on a spike-latency encoded Fashion-MNIST dataset using the surrogate gradient training algorithm \cite{xiaoFashionMNISTNovelImage2017, neftciSurrogateGradientLearning2019}.

Furthermore, additional tests were conducted utilizing a dataset designed specifically for the spiking domain, the SHD dataset \cite{cramerHeidelbergSpikingData2022}. This dataset creates biologically realistic spiking patterns by processing audio recordings through a cochlear model to create 700 spike trains for each individual audio recording. In total, there are 20 different classes corresponding to audio recordings of the numbers “zero” through “nine” in English and German (Supplementary Fig. 1). Again, a range of weight initializations was used to simulate a range of initial hidden layer firing rates.

\subsection{Noisy Training}
Neuromorphic hardware, particularly based on emerging memristive devices, are known to have noisy weight updates\cite{wangResistiveSwitchingMaterials2020, adamChallengesHinderingMemristive2018}. While significant progress in algorithmic solutions has been made to support highly precise programmability \cite{songProgrammingMemristorArrays2024, alibartHighPrecisionTuning2012} these come at the cost of additional hardware overhead. Utilizing the most successful weight initialization from earlier trials, $\sigma_{init}=0.001$, additional trials were performed with a noisy weight update. The noise levels investigated ranged from 1\% to 100\% of $\sigma_{init}$ which mirrored the experimental range of noise within oxide-based resistive switches \cite{garbinVariabilityStudyPCM2015}. While other work has proposed randomizations to STDP methods \cite{wangStochasticApproachSTDP2016}, we add noise to the backpropagation updates. At each batch weight update, a Gaussian noise was added to the ideal weight update (Fig. 2a). The noise level was determined by the standard deviation of the distribution. A range of noise levels were used to span a wide enough range to visualize progressive decline in accuracy across all networks. The noise levels are calculated as a ratio of the weight initialization from the input to hidden layer.

\subsection{Metrics and Analysis}
Neuroscience has produced a variety of metrics to analyze the spiking behavior of neurons at the population, pair, and individual level \cite{kassAnalysisNeuralData2014, gerstnerNeuronalDynamicsSingle2014}. Taking inspiration from neuroscience, we analyzed our networks using a few metrics suggested in the neuroscience literature and applied them in this artificial intelligence training environment. These neuroscience methods are further broken into two categories, temporal and frequency-based analysis influenced by the temporal and frequency encoding of the datasets. Additionally, we included standard metrics from machine learning analysis to provide a wider analysis of the network behavior. See Table 1 for an overview of the metrics.

\subsubsection{Temporal Metrics}
The temporal analysis of the network focused on the relative timing of the hidden layer neurons \cite{carianiTemporalCodingPeriodicity1999, carianiTemporalCodesComputations2004}. This was done by binning the number of spikes temporally across each simulation which corresponds to a single test case. Binning was done on a time step basis (1 ms). Across multiple simulations, we averaged the activity of the hidden layer across each training case. Activity was then compared across data classes and compared between excitatory and inhibitory neurons of the hidden layer.

To further analyze the relationship between neurons and their firing patterns we utilize the Van Rossum distance \cite{vanrossumNovelSpikeDistance2001}. This calculation is a single metric to compare spike trains between any neuron pairs using the previously marked spike times. Previously, the metric was used as a basis for other training methods \cite{zenkeSuperSpikeSupervisedLearning2018}, but we will use it here as a method of analysis of the network dynamics within the surrogate gradient training. The Van Rossum distance was calculated and averaged across the entire training data. Distance was calculated using an exponential decay kernel function, which mirrors the leaky-integrator output neurons and functions using the following equations.

\begin{equation}
f_{a}(t) = \sum_{i}^{N_{a}}{\delta(t_i)}
\end{equation}

\begin{equation}
g_{a}(t) = e^{\frac{-t}{\tau}}*f_{a}(t)
\end{equation}

\begin{equation}
dist(a,b) = \sqrt{\frac{1}{\tau}\int_{0}^{\infty}{(g_a(t)-g_b(t))^2}}
\end{equation}

In these equations, the spike train of neuron $a$ is given by $f_a(t)$ and is convolved with the exponential decay function to give $g_a(t)$ using $\tau_d$ = 1 ms. The distance is then the integral of the difference squared calculated as $dist(a,b)$. Based on this, the Van Rossum distance provides a numerical distance between neurons that mirrors the error calculation used for the network training.

\subsubsection{Frequency Metrics}
Frequency analysis methods focus on the inter-spike intervals (ISIs) of the networks \cite{carianiTemporalCodingPeriodicity1999, kreuzMeasuringSpikeTrain2007}. ISIs measure the time difference between successive spikes from any given neuron. While temporally encoded Fashion-MNIST doesn’t provide ISIs within the input layer due to the single spike per neuron, the hidden layer does have the ability to produce multiple spikes from a single neuron. ISIs were measured by simulating each training case and collecting the list of all hidden layer neuron ISIs across the entire dataset. This can be subdivided based on ISIs from excitatory or inhibitory neurons and based on dataset class. From the list of ISIs we can create a distribution and compare distributions relative to network architectures and across training.

The inverse of the ISIs was also computed to find the firing frequency and analyzed over the network training \cite{carianiTemporalCodesComputations2004, dayanTheoreticalNeuroscienceComputational2001}. Firing frequency is a known and often calculated metric that can be averaged over both the entire training data but can also be calculated over time to find the changing firing rate throughout a given simulation.

\subsubsection{Training Metrics}
Finally, the network itself can be analyzed in terms of its ability to train, and the weight modifications needed for the network to train \cite{rossbroichFluctuationdrivenInitializationSpiking2022}. Weights are the only parameter being updated by the training rule, and thus are fundamental in analyzing how the network is training. Additionally, weight analysis can allow us to see what features the network is learning throughout training. To compare network abilities in the classification task, the accuracy and loss values were calculated throughout the training indicating the speed and peak abilities of the network with both datasets \cite{pfeifferDeepLearningSpiking2018}. This metric allowed us to see the stability in network accuracy during the end of training, as well as the rate at which different networks were able to reach their final accuracy.

\begin{table}[!ht]
  \centering
  \caption{Summary of metrics.}
  \label{tab:metrics}
  \begin{tabular}{|p{3cm}|p{1.5cm}|p{3cm}|p{3cm}|}
    \hline
    \textbf{Metric} & \textbf{Type} & \textbf{Definition} & \textbf{Citations} \\
    \hline
    Spike Activity Over Time & Temporal & Binning and summing the number of spikes over duration of the trial. & Cariani 1999, wu et al. 2022 \cite{carianiTemporalCodingPeriodicity1999, wuTrainingSpikingNeural2022} \\
    \hline
    Van Rossum Distance & Temporal & Two spike trains are convolved with a kernel function, an exponential decay, and the squared difference is integrated over the entire trial. & Van Rossum 2001, Zenke and Ganguli 2018 \cite{vanrossumNovelSpikeDistance2001, zenkeSuperSpikeSupervisedLearning2018}\\
    \hline
    InterSpike Interval (ISI) & Frequency & Time interval between two subsequent spikes coming from the same neuron. & Cariani 1999, Kreuz et al. 2007 \cite{carianiTemporalCodingPeriodicity1999, kreuzMeasuringSpikeTrain2007}\\
    \hline
    Firing Frequency & Frequency & Inverse of interspike intervals, showing the number of spikes per time period. & Dayan and Abbott 2001, Kopsick et al. 2023 \cite{dayanTheoreticalNeuroscienceComputational2001, kopsickRobustRestingStateDynamics2023}\\
    \hline
    Accuracy and Loss & Training & Accuracy is the percentage of correct classifications of the testing data, loss is calculated as the negative-log-likelihood loss function for all training data. & Pfeiffer and Pfeil 2018, Bittar and Garner 2022, Yin et al. 2021 \cite{pfeifferDeepLearningSpiking2018, bittarSurrogateGradientSpiking2022, yinAccurateEfficientTimedomain2021}\\
    \hline
    Weight Distributions & Training & Connections between neurons have a weight scaling factor applied to spikes propagating through that synapse which change over training. & Rossbroich et al. 2022, Narkhede et al.  2022 \cite{rossbroichFluctuationdrivenInitializationSpiking2022, narkhedeReviewWeightInitialization2022}\\
    \hline
  \end{tabular}
\end{table}

\begin{table}[!ht]
    \caption{Average accuracy and Kruskal-Wallis p-value from Fig. 5a-c.}
    \begin{tabular}{|p{1.5cm}|p{1cm}|p{1cm}|p{1cm}|p{1cm}|p{1cm}|p{1cm}|p{1cm}|p{1cm}|p{1cm}|}
    \hline
     Fashion-MNIST & \multicolumn{3}{c|}{All Pre-Training} & \multicolumn{3}{c|}{Successful Post-Training} & \multicolumn{3}{c|}{Failed Post-Training} \\
     \hline
     Ratio & Average Accuracy & Median E-E, E-I, I-I Distance & P-value (N) & Average Accuracy & Median E-E, E-I, I-I Distance & P-value (N) & Average Accuracy & Median E-E, E-I, I-I Distance & P-value (N) \\
     \hline
     50:50 & 9.83\% & 1.291, 1.283, 1.278 & 0.9920 (141) & 68.92\% & 0.773, 0.992, 1.079 & 0.0091 (32)            & 15.34\% & 1.457, 1.540, 1.446 & 0.2260 (109) \\
     \hline
80:20 & 9.83\% & 1.297, 1.289, 1.287 & 0.9954 (145) & 80.71\% & 0.890, 1.025, 1.100 & \textless{}0.0001 (41) & 15.03\% & 1.532, 1.626, 1.553 & 0.0880 (104) \\
    \hline
95:5  & 9.86\% & 1.284, 1.272, 1.273 & 0.9979 (146) & 80.86\% & 0.938, 1.020, 1.108 & \textless{}0.0001 (53) & 13.29\% & 1.539, 1.769, 1.630 & 0.0018 (93) \\
    \hline
    \end{tabular}
\end{table}

\begin{table}[!ht]
    \caption{Average accuracy and Kruskal-Wallis p-value from Fig. 5d-f.}
    \begin{tabular}{|p{1.5cm}|p{1cm}|p{1cm}|p{1cm}|p{1cm}|p{1cm}|p{1cm}|p{1cm}|p{1cm}|p{1cm}|}
    \hline
     SHD & \multicolumn{3}{c|}{All Pre-Training} & \multicolumn{3}{c|}{Successful Post-Training} & \multicolumn{3}{c|}{Failed Post-Training} \\
     \hline
     Ratio & Average Accuracy & Median E-E, E-I, I-I Distance & P-value (N) & Average Accuracy & Median E-E, E-I, I-I Distance & P-value (N) & Average Accuracy & Median E-E, E-I, I-I Distance & P-value (N) \\
     \hline
    50:50 & 5.00\% & 1.181, 1.185, 1.175 & 0.9859 (137)      & 37.47\%     & 2.178, 2.357, 2.380 & 0.0087 (39)                   & 10.54\%     & 1.339, 1.310, 1.366 & 0.7517 (98)    \\
    \hline
80:20 & 4.90\%        & 1.187, 1.189, 1.195                                               & 0.9998 (137)                  & 37.76\%     & 1.929, 2.359, 2.444 & \textless{}0.0001 (45)        & 10.75\%     & 1.285, 1.389, 1.356 & 0.4166 (92)   \\
    \hline
95:5  & 4.95\%        & 1.179, 1.167, 1.138                                               & 0.9040 (140)                  & 39.99\%     & 1.862, 2.775, 2.622 & \textless{}0.0001 (30)        & 7.97\%      & 1.265, 1.360, 1.319 & 0.0638 (110)  \\   
    \hline
    \end{tabular}
\end{table}

\newpage
\backmatter

\bmhead{Acknowledgements}

The authors acknowledge the use of high-performance computing clusters, advanced support from the research technology services, and IT support at The George Washington University.

This work was supported by the Department of Energy Office of Science, CRCNS collaborative effort under grant numbers DE-SC00023000 (GWU) and DE-SC0022998 (GMU). Additional support was provided by the National Institute of Health under grant number R01NS39600 and Air Force Office of Scientific Research under grant number FA9550-23-1-0173.

\vspace{10mm}

\bibliography{sn-article}

\end{document}